\pdfoutput=1

\documentclass[11pt]{article}

\usepackage{acl}

\usepackage{times}
\usepackage{latexsym}

\usepackage[T1]{fontenc}

\usepackage[utf8]{inputenc}

\usepackage{microtype}

\usepackage{inconsolata}

\usepackage{algorithm}
\usepackage{algpseudocode}
\usepackage{multicol}
\usepackage{multirow}
\usepackage{amsmath}
\usepackage{amsfonts}
\usepackage{ulem}
\usepackage{svg}
\usepackage{tabularx}
\usepackage{makecell}
\usepackage{stfloats}
\usepackage{arydshln}
\usepackage{adjustbox}
\usepackage{makecell}
\usepackage{array}

\usepackage{booktabs}
\usepackage{caption}
\usepackage{subcaption}
\usepackage{tikz}
\usepackage{pgfplots}
\usepackage{tikz-dependency}
\usepackage{tikz-qtree}
\usepgflibrary{arrows.meta}
\usepgfplotslibrary{fillbetween}
\usetikzlibrary{fit, calc, decorations.pathreplacing, positioning, shapes.geometric, backgrounds}

\title{Ungrammatical-syntax-based In-context Example Selection \\for Grammatical Error Correction}

\author{
Chenming Tang \quad
Fanyi Qu \quad
Yunfang Wu\thanks{\ \ \ Corresponding author.} \\
  National Key Laboratory for Multimedia Information Processing, Peking University \\
  MOE Key Laboratory of Computational Linguistics, Peking University\\
  School of Computer Science, Peking University \\
  \texttt{\{tangchenming@stu, fanyiqu@, wuyf@\}pku.edu.cn}
}

\begin{document}

\maketitle

\begin{abstract}

In the era of large language models (LLMs), in-context learning (ICL) stands out as an effective prompting strategy that explores LLMs' potency across various tasks. However, applying LLMs to grammatical error correction (GEC) is still a challenging task. In this paper, we propose a novel ungrammatical-syntax-based in-context example selection strategy for GEC. Specifically, we measure similarity of sentences based on their syntactic structures with diverse algorithms, and identify optimal ICL examples sharing the most similar ill-formed syntax to the test input. Additionally, we carry out a two-stage process to further improve the quality of selection results. On benchmark English GEC datasets, empirical results show that our proposed ungrammatical-syntax-based strategies outperform commonly-used word-matching or semantics-based methods with multiple LLMs. This indicates that for a syntax-oriented task like GEC, paying more attention to syntactic information can effectively boost LLMs' performance. Our code will be publicly available after the publication of this paper.

\end{abstract}

\section{Introduction}

\begin{table*}[thbp]
  \centering
  \small
    \scalebox{0.95}{
    \begin{tabular}{cll}
    \midrule
          & \textbf{Source (Erroneous Sentence)} & \textbf{Target (Corrected Sentence)} \\
    \midrule
    Input & No smoking in 
    \textit{the} public places. & No smoking in public places. \\
    \midrule
    \multirow{2}{*}{BM25}  & I am writing to complain about the suggested & I am writing to complain about the suggested \\
    & 
    \textit{bar}
    on smoking in public areas. & 
    \textit{ban} on smoking in public areas. \\
    Poly. & No future for 
    \textit{the} public transport? & No future for public transport? \\
    \midrule
    \end{tabular}
    }
  \caption{An example comparing the selection results of BM25 and Polynomial Distance ("Poly." in the table).}
  \label{tab:introcase}%
\end{table*}%

Recently, large language models (LLMs) have shown awesome power in many areas and ended the contest on many tasks \cite{chowdhery2023palm, bubeck2023sparks, touvron2023llama}. Unfortunately for LLMs, grammatical error correction (GEC), which aims at automatically correcting grammatical errors in erroneous text \citep{bryant2023grammatical}, is still a challenging task where they cannot beat conventional 
models yet. \citet{fang2023chatgpt} and \citet{loem-etal-2023-exploring} 
explore the performance of LLMs on GEC, 
demonstrating mainstream LLMs lag over 10 points behind the state-of-the-art result. 
Therefore, it is significant to explore 
new strategies 
to further improve the
power of LLMs on GEC.

As a helpful prompting strategy, 
in-context learning (ICL) has achieved impressive results
on many tasks \cite{dong2022survey, min2022rethinking}. 
In ICL, several in-context examples are 
presented to LLMs as demonstrations before the 
input test sample in order to make LLMs aware of the requirement and output format of the specific task, thereby enhancing LLMs' performance during the subsequent generation process.
Since the quality of in-context examples plays a crucial role under the few-shot setting, 
some 
strategies of 
example selection have been proposed \cite{agrawal-etal-2023-context,li-etal-2023-unified,pmlr-v202-ye23c,gupta-etal-2023-coverage}.

To the best of our knowledge, no existing work on ICL example selection has taken syntactic information into consideration. 
However, GEC aims to correct grammatical errors and is a typical syntax-oriented task. 
In GEC, common errors can be classified into four types: \textit{misuse}, \textit{missing}, \textit{redundancy} and \textit{word order} (\citealp{bryant-etal-2017-automatic}; \citealp{zhang-etal-2022-mucgec}), and the last three of which are closely related to syntactic structure. 
That is, the \textit{missing}, \textit{redundancy} or \textit{disorder} of text constituents 
may lead to 
ill-formed syntax \citep{zhang-etal-2022-syngec}, 
suggesting the important role syntax plays in GEC. Hence, selecting in-context examples based on syntactic 
structure is likely to benefit LLMs more than conventional 
word-matching-based or semantics-based approaches. 

Comparing with semantic similarity, syntactic similarity of text is less-studied. Previous works have leveraged the similarity of dependency trees to help multi-document summarization ~\cite{ozates-etal-2016-sentence} and semantic textual similarity~\cite{le-etal-2018-acv}. To compute syntactic similarity, 
several effective algorithms computing similarity between syntactic trees have been proposed. Tree Kernel is a typical one, which counts the shared sub-structures of two trees to measure their similarity (\citealp{collins-duffy-2002-new}; \citealp{vishwanathan2004fast}; \citealp{moschitti-2006-efficient}). Polynomial Distance is another handy one, which converts syntactic trees into polynomials 
and then computes the distances 
\citep{liu2022quantifying}.


In this paper, we propose a novel ICL example selection strategy for GEC, by computing similarities of syntactic trees 
on ungrammatical sentences.
Specially, we apply the syntactic similarity algorithms (Tree Kernel and Polynomial Distance) 
to dependency trees generated by a GEC-oriented parser (GOPar) proposed by \citet{zhang-etal-2022-syngec}, which is more reliable and provides error information when parsing ungrammatical sentences.
Moreover, we carry out a two-stage process.
In the first stage, namely \textit{selection}, a fast and general method like BM25 
is applied to filter out most of the 
irrelevant instances from the training data and obtain a much smaller candidate set. In the second stage, namely \textit{ranking}, 
the more powerful syntax-based method is 
implemented to 
find out the best $k$ instances 
as the final in-context examples.

To give a quick view of the superiority of our method, Table \ref{tab:introcase} shows an example illustrating the difference between 
BM25 selection and our ungrammatical-syntax-based method with Polynoimal Distance selection. BM25 
only selects examples with similar words while Polynomial Distance 
is able to select those with similar grammatical errors, which will benefit the GEC task more. 


We conduct experiments on two mainstream English GEC datasets, BEA-2019 ~\citep{bryant-etal-2019-bea} and CoNLL-2014~\citep{ng-etal-2014-conll}. 
According to experimental results, Polynomial Distance and its weighted version 
achieve competitive results even under the single-stage setting, improving the performance by around 3 points and 2 points on BEA-19 and CoNLL-14 respectively. With the help of our two-stage selection, Tree Kernel gets its power unlocked and Polynomial Distance also benefits, leading to a further 1-point and 0.4-point improvement on BEA-19 and CoNLL-14 respectively. 
Overall, our ungrammatical-syntax-based in-context example selection methods secure the best results under all settings, outperforming conventional baselines by a margin of nearly 3 F$_{0.5}$ points on average.

Our contributions can be summarized as follows:

\begin{itemize}
    \item We propose a novel ICL example selection method based on ungrammatical syntactic similarity to improve LLMs' performance on GEC. 
    To the best of our knowledge, this is the first time that
    knowledge of syntactic structure is introduced
    to 
    ICL example selection for GEC.
    \item We explore a two-stage selection strategy on GEC, where 
    superficial word-similarity-based or semantics-based methods are used in the first stage and deep syntax-similarity-based ones are used in the second stage. It further improves LLMs' performance and achieves competitive results. 
    \item We want to re-draw the natural language processing (NLP) community's attention to the significance of syntactic information. 
    In this work, we show that syntax-related knowledge 
    helps LLMs correct grammatical errors better. 
    We believe our 
    methods can 
    be smoothly transferred to many other syntax-related tasks, like machine translation (MT) and information extraction (IE).
\end{itemize}

\begin{figure*}[htbp]
    \centering

    \begin{adjustbox}{center}
    \includegraphics[scale=0.4]{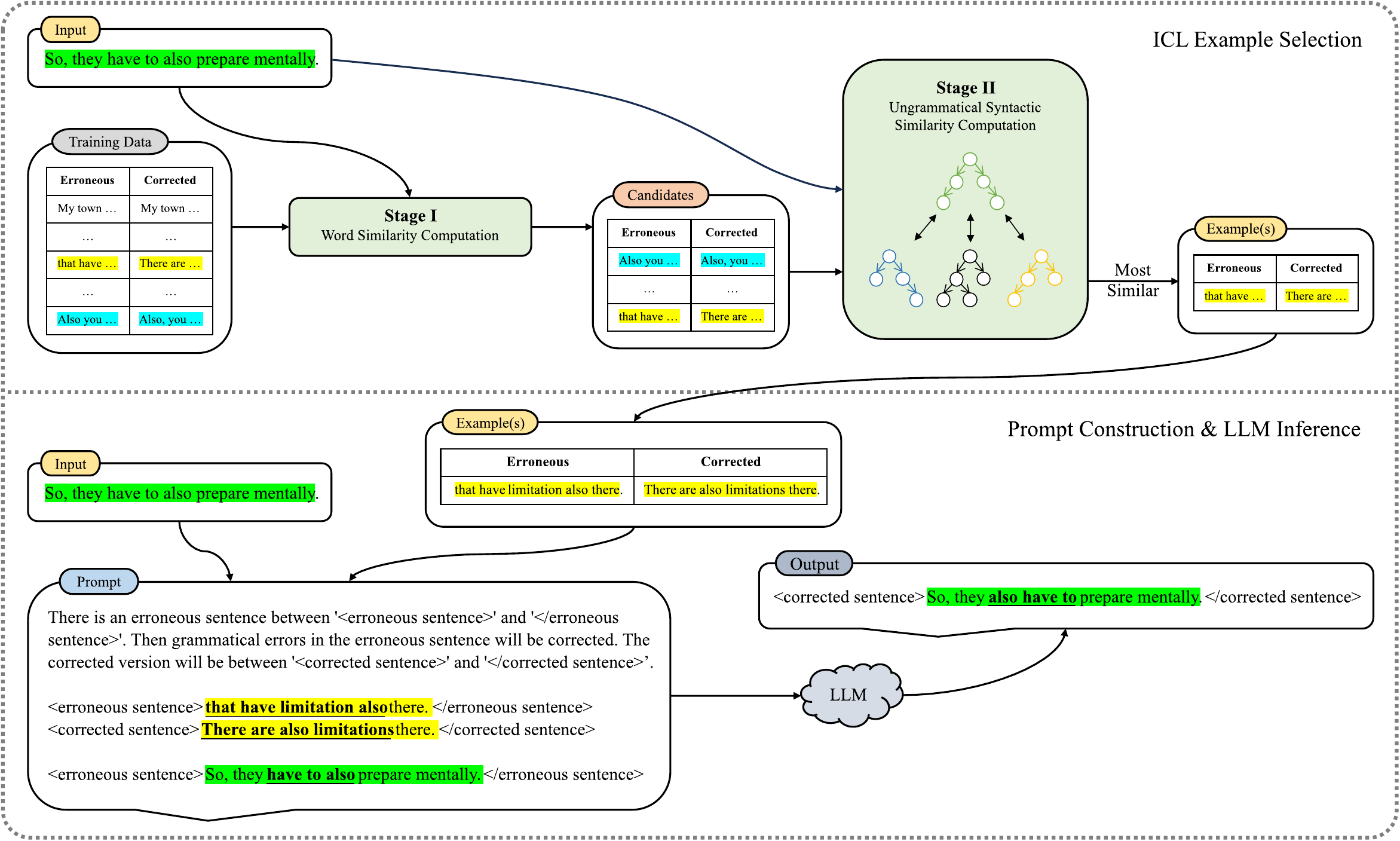}

    \end{adjustbox}
    
    \caption{Our two-stage selection and ICL workflow. For each input test sample, Stage \textbf{I} computes word similarities with BM25 or BERT representation between the input and all training data and select the top-$1000$ as candidates. Then, Stage \textbf{II} computes ungrammatical syntactic similarities with tree kernel or polynomial distance between the input and candidates to select the most similar $k$ example(s). After that, we concatenate the input after the $k$ examples to construct the prompt for LLM inference. In the end, the LLM outputs the final result.}
    \label{fig:icl}
\end{figure*}

\section{Related Work}
\subsection{Grammatical Error Correction}
In the past few years, the GEC task has been dominated by sequence-to-sequence machine translation models \cite{junczys-dowmunt-etal-2018-approaching, rothe-etal-2021-simple} and sequence-to-edit tagging models \cite{omelianchuk-etal-2020-gector, tarnavskyi-etal-2022-ensembling}, both based on Transformer \citep{vaswani2023attention}.

Nowadays, with the finalization of mainstream models, further explorations on GEC mainly focus on two aspects. For one thing, injecting all kinds of additional knowledge into GEC models has proved helpful. The additional knowledge can be part-of-speech (POS) \citep{wu-wu-2022-spelling}, syntax tree \citep{zhang-etal-2022-syngec}, speech representation \citep{fang-etal-2023-improving}, abstract meaning representation (AMR) \citep{cao-zhao-2023-leveraging}, error type \citep{yang-etal-2023-leveraging}, etc. For another, multi-stage strategies help refine models' predictions. The multi-stage workflow can be permutation \& decoding \citep{yakovlev-etal-2023-gec}, detection \& correction \citep{li-etal-2023-templategec}, re-ranking \citep{zhang-etal-2023-bidirectional}, etc.

With the rising of powerful large language models (LLMs), some works have begun exploring their performance on GEC (\citealp{loem-etal-2023-exploring}; \citealp{fang2023chatgpt}), showing that LLMs cannot beat conventional models on GEC yet.

\subsection{Syntactic Similarity}

In computational linguistics (CL), previous works compared syntax trees of different languages to measure their similarities (\citealp{oya-2020-syntactic}; \citealp{liu2022quantifying}). In NLP, most works on text similarity focus on the semantic perspective (\citealp{gomaa2013survey}, \citealp{reimers-gurevych-2019-sentence}; \citealp{chandrasekaran2021evolution}), syntactic similarity of text is less-studied. \citet{ozates-etal-2016-sentence} used similarity of dependency trees to help multi-document summarization. \citet{le-etal-2018-acv} proposed ACV-tree (Attention Constituency Vector-tree), which combines word weight, word representation and constituency tree, to help the task of semantic textual similarity.

Syntactic similarity is usually represented by similarity between syntax trees. Tree similarity can be measured by various algorithms including Edit Distance \citep{Reis2004AutomaticWN}, Polynomial Distance \citep{liu2022quantifying}, Subset Tree Kernel (SSTK) \citep{collins-duffy-2002-new}, SubTree Kernel (STK) \citep{vishwanathan2004fast}, and Partial Tree Kernel (PTK) \citep{moschitti-2006-efficient}.

\subsection{Large Language Models and In-context Learning}

In recent years, LLMs have shown their awesome power in many areas (\citealp{brown2020language}; \citealp{chowdhery2023palm}). Due to the limitation of computing resources,
the focus of research on LLMs turns to the inference stage, trying to exploit the potency of LLMs with inference-only strategies.

ICL is a successful inference strategy that can make LLMs perform as well as fine-tuned models on many tasks (\citealp{brown2020language}; \citealp{von2023transformers}), where several in-context examples are given to LLMs as demonstrations before the actual test input. Instead of randomly sampling examples from the training set, recent works have boosted the performance of ICL by selecting in-context examples using various strategies. \citet{agrawal-etal-2023-context} proposed R-BM25, a word overlap and coverage based selection strategy for machine translation.
\citet{li-etal-2023-unified} proposed training a Unified Demonstration Retriever (UDR) for ICL on a wide range of tasks.
\citet{pmlr-v202-ye23c} treated in-context examples as a whole and selected examples on a subset level with the help of Determinantal Point Processes (DPPs).
\citet{gupta-etal-2023-coverage} also selected examples as an entire set, with contextual-embedding-level coverage as the goal.

Besides normal ICL, Chain-of-thought (CoT) (\citealp{wei-etal-2022-chain}; \citealp{kojima2023large}) is another effective inference strategy in current favor, where LLMs are prompted to think step by step and answer with intermediate rationales.

\section{Preliminaries}

\subsection{Syntax Parser for Ungrammatical Sentences}
\label{subsec:gopar}

\definecolor{brickred}{HTML}{b92622}
\definecolor{midnightblue}{HTML}{005c7f}
\definecolor{salmon}{HTML}{f1958d}
\definecolor{burntorange}{HTML}{f19249}
\definecolor{junglegreen}{HTML}{4dae9d}
\definecolor{forestgreen}{HTML}{499c5e}
\definecolor{pinegreen}{HTML}{3d8a75}
\definecolor{seagreen}{HTML}{6bc1a2}
\definecolor{limegreen}{HTML}{97c65a}
\newcommand{\white}[1]{\textcolor{white}{#1}}
\newcommand{\brickred}[1]{\textcolor{brickred}{#1}}
\newcommand{\midnightblue}[1]{\textcolor{midnightblue}{#1}}
\newcommand{\salmon}[1]{\textcolor{salmon}{#1}}
\newcommand{\junglegreen}[1]{\textcolor{junglegreen}{#1}}
\newcommand{\forestgreen}[1]{\textcolor{forestgreen}{#1}}
\newcommand{\pinegreen}[1]{\textcolor{pinegreen}{#1}}
\newcommand{\seagreen}[1]{\textcolor{seagreen}{#1}}

\begin{figure*}[th!]
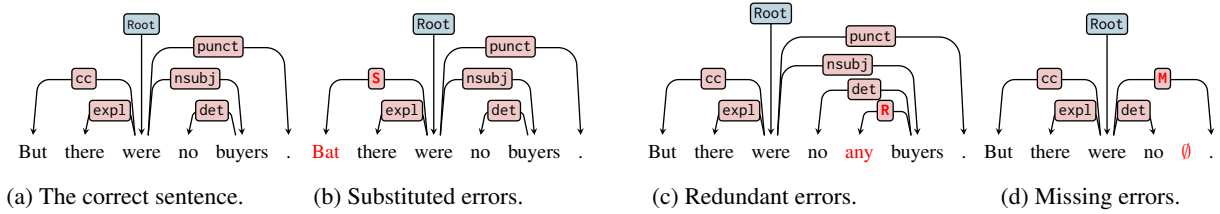

  \begin{subfigure}[b]{0.2\textwidth}
    \centering
    \scalebox{0.62}{
        \begin{dependency}
          \begin{deptext}[column sep=0.2cm,font=\small]
            \large{But} \& \large{there} \& \large{were} \& \large{no} \& \large{buyers} \& \large{.} \\ 
          \end{deptext}
          \Large{
          \deproot[edge vertical padding=0.6ex, edge height=12ex, label style={fill=midnightblue!25, thick}, edge style={thick}]{3}{\texttt{\large{Root}}}
          \depedge[edge vertical padding=0.6ex, edge height=5.2ex, label style={fill=brickred!25, thick}, thick]{3}{1}{\texttt{cc}}
          \depedge[edge vertical padding=0.6ex, edge height=2.2ex, label style={fill=brickred!25, thick}, thick]{3}{2}{\texttt{expl}}
            \depedge[edge vertical padding=0.6ex, edge height=5.2ex, label style={fill=brickred!25, thick}, thick]{3}{5}{\texttt{nsubj}}
          \depedge[edge vertical padding=0.6ex, edge height=8.2ex, label style={fill=brickred!25, thick}, thick]{3}{6}{\texttt{punct}}
          \depedge[edge vertical padding=0.6ex, edge height=2.2ex, label style={fill=brickred!25, thick}, thick]{5}{4}{\texttt{det}}}
    
        \end{dependency}
    }
    
    \caption{The correct sentence.}
    \label{fig:origin-sent}
  \end{subfigure}
  \hfill
   \begin{subfigure}[b]{0.2\textwidth}
    \centering
    \scalebox{0.62}{
    
    \begin{dependency}
      \begin{deptext}[column sep=0.2cm,font=\small]
        \large{\textcolor{red}{Bat}} \& \large{there} \& \large{were} \& \large{no} \& \large{buyers} \& \large{.} \\ 
      \end{deptext}
      \Large{
      \deproot[edge vertical padding=0.6ex, edge height=12ex, label style={fill=midnightblue!25, thick}, edge style={thick}]{3}{\texttt{Root}}
      \depedge[edge vertical padding=0.6ex, edge height=5.2ex, label style={fill=brickred!25, thick}, thick]{3}{1}{\textcolor{red}{\texttt{\textbf{S}}}}
      \depedge[edge vertical padding=0.6ex, edge height=2.2ex, label style={fill=brickred!25, thick}, thick]{3}{2}{\texttt{expl}}
        \depedge[edge vertical padding=0.6ex, edge height=5.2ex, label style={fill=brickred!25, thick}, thick]{3}{5}{\texttt{nsubj}}
      \depedge[edge vertical padding=0.6ex, edge height=8.2ex, label style={fill=brickred!25, thick}, thick]{3}{6}{\texttt{punct}}
      \depedge[edge vertical padding=0.6ex, edge height=2.2ex, label style={fill=brickred!25, thick}, thick]{5}{4}{\texttt{det}}
      }

    \end{dependency}
    }
    \caption{Substituted errors.}
    \label{fig:s-error}
  \end{subfigure}
  \hfill
  \hfill
   \begin{subfigure}[b]{0.2\textwidth}
    \centering
    \scalebox{0.62}{
    \begin{dependency}
      \begin{deptext}[column sep=0.2cm,font=\small]
        \large{But} \& \large{there} \& \large{were} \& \large{no} \& \textcolor{red}{\large{any}} \& \large{buyers} \& \large{.} \\ 
      \end{deptext}
      \Large{
      \deproot[edge vertical padding=0.6ex, edge height=13ex, label style={fill=midnightblue!25, thick}, edge style={thick}]{3}{\texttt{Root}}
      \depedge[edge vertical padding=0.6ex, edge height=5.2ex, label style={fill=brickred!25, thick}, thick]{3}{1}{\texttt{cc}}
      \depedge[edge vertical padding=0.6ex, edge height=2.2ex, label style={fill=brickred!25, thick}, thick]{3}{2}{\texttt{expl}}
        \depedge[edge vertical padding=0.6ex, edge height=6.5ex, label style={fill=brickred!25, thick}, thick]{3}{6}{\texttt{nsubj}}
      \depedge[edge vertical padding=0.6ex, edge height=9.2ex, label style={fill=brickred!25, thick}, thick]{3}{7}{\texttt{punct}}
      \depedge[edge vertical padding=0.6ex, edge height=4.2ex, label style={fill=brickred!25, thick}, thick]{6}{4}{\texttt{det}}
      \depedge[edge vertical padding=0.6ex, edge height=2.2ex, label style={fill=brickred!25, thick}, thick]{6}{5}{\textbf{\textcolor{red}{\texttt{R}}}}
        }
    \end{dependency}
    }
    \caption{Redundant errors.}
    \label{fig:r-error}
  \end{subfigure}
  \hfill
  \hfill
   \begin{subfigure}[b]{0.2\textwidth}
    \centering
    \scalebox{0.62}{
    \begin{dependency}
      \begin{deptext}[column sep=0.2cm,font=\small]
        \large{But} \& \large{there} \& \large{were} \& \large{no} \& \textcolor{red}{\large{$\emptyset$}} \& \large{.} \\ 
      \end{deptext}
      \Large{
      \deproot[edge vertical padding=0.6ex, edge height=12ex, label style={fill=midnightblue!25, thick}, edge style={thick}]{3}{\texttt{Root}}
      \depedge[edge vertical padding=0.6ex, edge height=5.2ex, label style={fill=brickred!25, thick}, thick]{3}{1}{\texttt{cc}}
      \depedge[edge vertical padding=0.6ex, edge height=2.2ex, label style={fill=brickred!25, thick}, thick]{3}{2}{\texttt{expl}}
      \depedge[edge vertical padding=0.6ex, edge height=5.2ex, label style={fill=brickred!25, thick}, thick]{3}{6}{\textcolor{red}{\texttt{\textbf{M}}}}
      \depedge[edge vertical padding=0.6ex, edge height=2.2ex, label style={fill=brickred!25, thick}, thick]{3}{4}{\texttt{det}}
        }
    \end{dependency}
    }
    \caption{Missing errors.}
    \label{fig:m-error}
  \end{subfigure}
  \hfill
  \caption{Original illustration of GOPar from \citet{zhang-etal-2022-syngec}. $\emptyset$ denotes the missing word.}
  \label{fig:scheme-three-error-types}
\end{figure*}
Unlike most NLP tasks, which take correct sentences as input, the GEC task considers erroneous text as input. This gives rise to an issue that mainstream parsers may fail to obtain the expected dependency tree for the erroneous text.   

To solve this problem, \citet{zhang-etal-2022-syngec} built a tailored GEC-Oriented dependency Parser (GOPar) 
based on parallel GEC training data,
which is much more reliable when handling ungrammatical sentences than conventional parsers. Concretely, GOPar sets "S" (Substituted), "R" (Redundant) or "M" (Missing) labels 
to deal with different kinds of grammatical errors in the sentence, which inject additional information of 
errors into the 
dependency tree. Figure \ref{fig:scheme-three-error-types} shows the original illustration of GOPar from \citet{zhang-etal-2022-syngec}. 

Most previous works 
computing syntactic similarity
were based on grammatical sentences 
with standard parsing trees~\cite{ozates-etal-2016-sentence,oya-2020-syntactic}. However, 
in GEC, we only have the ungrammatical source sentences, on which conventional parsers may perform poorly. So we apply 
the algorithms of tree similarity on the parsing results of GOPar, to compute syntactic similarities between test sample and training instances. 
We follow the official guidance of SynGEC~\footnote{https://github.com/HillZhang1999/SynGEC} to run GOPar. We use \texttt{biaffine-dep-electra-en-gopar} provided by SynGEC as the model for parsing.

\subsection{Syntactic Similarity with Tree Kernel}
We follow the unified Tree Kernel method proposed by 
\citet{moschitti-2006-efficient}, 
which can compute kernels of subset trees 
defined by \citet{collins-duffy-2002-new}, subtrees 
defined by \citet{vishwanathan2004fast} and partial trees 
defined in their own work.

For simplicity, we imitate the algorithm described in \citet{le-etal-2018-acv} and 
design the following algorithm (shown in Algorithm \ref{alg:ntk}) to 
implement a simple 
version of Tree Kernel.

\begin{algorithm}
\small
\caption{Similarity with Tree Kernel}\label{alg:ntk}
\begin{algorithmic}
\Procedure{CompSim}{$N_1, N_2$}

\State $K \gets 0$

\For {each node $n_i$ in $N_1$}
\For {each node $n_j$ in $N_2$}
\If{$n_i.label = n_j.label$}
\If{$n_i$ and \ $n_j$ are both leaves}
\State $K \gets K + 1$
\ElsIf{$n_i$ and \ $n_j$ are both non-leaves}
\State $K \gets K\ + $ \textsc{CompSim}($n_i$, $n_j$)
\EndIf
\EndIf
\EndFor
\EndFor

\State $K \gets K / (N_1.size \times N_2.size)$
\State return $K$
\EndProcedure
\end{algorithmic}
\end{algorithm}

For two trees $T_1$ and $T_2$, we conduct \textsc{CompSim} between their root nodes $N_1$ and $N_2$ to get a similarity score.

\subsection{Syntactic Similarity with Polynomial Distance}
\citet{liu2022quantifying} converted trees into polynomials and took the distances between polynomials as tree distances to measure syntactic similarities of dependency trees.

Given the number of dependency labels $d$, the dependency trees will be represented into polynomials recursively on two variable set: $X=\{x_1, x_2...x_d\}$ and $Y=\{y_1, y_2, ...y_d\}$. In the dependency tree, for each leaf $n^l$ with label $l$, the corresponding polynomial is $P(n^l, X, Y) = x_l$. Then, for each non-leaf $m_l$ with label $l$, the corresponding polynomial is $P(m^l, X, Y) = y_l + \prod_{i=1}^{k}{P(n_i, X, Y)}$, where $n_1, ..., n_k$ are all child nodes of $m_l$. In this way, the polynomial of the root node is regarded as the polynomial representation of a tree.

To compute similarity more conveniently,
for each term $cx_1^{e_{x_1}}x_2^{e_{x_2}}...x_d^{e_{x_d}}y_1^{e_{y_1}}y_2^{e_{y_2}}...y_d^{e_{y_d}}$ in the dependency polynomial, we 
write it as a term vector with $2d+1$ entries:
$$t = [e_{x_1}, e_{x_2}, ..., e_{x_d}, e_{y_1}, e_{y_2}, ..., e_{y_d}, c],$$
where each entry represents the exponent of the corresponding variable. In this way, a polynomial $P$ can be written as a set of term vectors $\mathcal{V}_P$. Then, we 
compute the distance between two polynomials as:

\begin{equation}
\label{eq:poly}
\small
d(P, Q) = \frac{\sum\limits_{s \in \mathcal{V}_P}{\min\limits_{t \in \mathcal{V}_Q} {\parallel s - t \parallel}_1} + \sum\limits_{t \in \mathcal{V}_Q}{\min\limits_{s \in \mathcal{V}_P} {\parallel s - t \parallel}_1}}{\mid \mathcal{V}_P \mid + \mid \mathcal{V}_Q \mid},
\end{equation}
where ${\parallel s - t \parallel}_1$ denotes the Manhattan distance \citep{Craw2017} between term vector $s$ and $t$.

\section{Methodology}

\subsection{In-context Learning Workflow for GEC}

\begin{table*}[htbp]
  \centering
  \scalebox{0.8}{
    \begin{tabularx}{\textwidth}{X|X}
    \hline
    \multicolumn{1}{c|}{\textbf{LLaMA-2}} & \multicolumn{1}{c}{\textbf{GPT-3.5}} \\
    \hline
    \small
    There is an erroneous sentence between '<erroneous sentence>' and '</erroneous sentence>'. Then grammatical errors in the erroneous sentence will be corrected. The corrected version will be between '<corrected sentence>' and '</corrected sentence>'. \newline <erroneous sentence> \{$e_1$\} </erroneous sentence> \newline <corrected sentence> \{$c_1$\} </corrected sentence> \newline <erroneous sentence> \{$e_2$\} </erroneous sentence> \newline <corrected sentence> \{$c_2$\} </corrected sentence> \newline <erroneous sentence> \{$e_3$\} </erroneous sentence> \newline <corrected sentence> \{$c_3$\} </corrected sentence> \newline <erroneous sentence> \{$e_4$\} </erroneous sentence> \newline <corrected sentence> \{$c_4$\} </corrected sentence> \newline <erroneous sentence> \{$e_{test}$\} </erroneous sentence> \newline <corrected sentence>    
    & 
    \small
    "system": You are a grammar correction assistant. The user will give you a sentence with grammatical errors (between '<erroneous sentence>' and '</erroneous sentence>'). You need to correct the sentence (between '<corrected sentence>' and '</corrected sentence>'). Requirements: 1. Make as few changes as possible. 2. Make sure the sentence has the same meaning as the original sentence. 3. If there is no error, just output 'No errors found'. \newline "user":  <erroneous sentence> \{$e_1$\} </erroneous sentence> \newline "assistant": <corrected sentence> \{$c_1$\} </corrected sentence> \newline "user":  <erroneous sentence> \{$e_2$\} </erroneous sentence> \newline "assistant": <corrected sentence> \{$c_2$\} </corrected sentence> \newline "user":  <erroneous sentence> \{$e_3$\} </erroneous sentence> \newline "assistant": <corrected sentence> \{$c_3$\} </corrected sentence> \newline "user":  <erroneous sentence> \{$e_4$\} </erroneous sentence> \newline "assistant": <corrected sentence> \{$c_4$\} </corrected sentence> \newline "user":  <erroneous sentence> \{$e_{test}$\} </erroneous sentence>
    \\
    \hline
    \end{tabularx}
    }
  \caption{Prompts we use. $e$ and $c$ denote the erroneous and corrected sentences of in-context examples or test samples respectively.}
  \label{tab:prompt}
\end{table*}

Based on LLMs, our ungrammatical-syntax-based example selection and few-shot ICL workflow is illustrated  
in Figure \ref{fig:icl}. Specially, when faced with a test input, we search through the training data to find the best example(s) for in-context learning. Then, 
both the source (erroneous) and the target (corrected) sentences 
of the example(s) are inserted into the prompt as demonstrations, with 
the test sample concatenated at the end. In this way, LLMs can learn the GEC task from the demonstrations and perform better correction 
on the test input. 
In this framework, a set of high-quality in-context examples are crucial to 
lead LLMs to a better performance. 
Prompts used in this work are shown in Table \ref{tab:prompt}.

\subsection{Ungrammatical-syntax-based Selection}
We parse all source sentences in both training and test data with GOPar introduced in Section \ref{subsec:gopar}. Then, for each test input, we search the training data to find most syntactically similar examples with the help of Tree Kernel or Polynomial Distance. These examples will serve as in-context demonstrations in the prompts shown in Table \ref{tab:prompt}.

\paragraph{Weighting Ungrammatical Nodes with Polynomial Distance}
We hypothesize that LLMs benefit more from similar grammatical errors, and error nodes with similar neighboring syntactic structure 
lead to similar error patterns. Therefore, assigning higher weights to ungrammatical nodes can select examples with error patterns closer to the test sample. Hence, besides the original Polynomial Distance algorithm, we also explore a weighted version. When computing the Manhattan distance between two term vectors, we assign a higher weight to entries corresponding to labels with error information ("S", "R" and "M"). In our experiment, as a preliminary attempt, we set the weight to 2.

\subsection{Two-stage Selection}
In previous works, a two-stage select-then-rank strategy  
performs well in in-context learning
(\citealp{wu-etal-2023-self}; \citealp{agrawal-etal-2023-context}). To be specific, a fast and general method
is used to filter out most of the not-so-relevant instances from training data and get a much smaller candidate set with high quality, which is called \textit{selection}. 
After that, a specific and powerful method 
is used to rank the instances in the candidate set and 
obtain the top-$k$ best training instances,
which is called \textit{ranking}. 
Motivated by this, we also design a two-stage sample selection mechanism for GEC. 
\paragraph{Stage 1: BM25/BERT Selection} 
First, we explore \textit{selection} with BM25 or BERT representation to obtain candidate examples, and the size of candidate set is 
1000 in our experiment.


\textbf{BM25}~\citep{Robertson1994OkapiAT} is a widely-used retrieval algorithm based on term frequency, inverse document frequency and length normalization. Many recent works 
regard BM25 as a strong baseline for in-context example selection 
(\citealp{agrawal-etal-2023-context}; \citealp{li-etal-2023-unified}). In our work, we take the 
input test sample as the query and source sentences of all training data as the document.

\textbf{BERT Representation}
~\citet{li-etal-2023-unified} make use of SentenceBERT \citep{reimers-gurevych-2019-sentence} to get sentence representations and then compared similarities of sentences. For simplicity, we adopt the more frequently-used BERT \citep{devlin-etal-2019-bert} instead. 
In our work, we take the BERT representation of the \texttt{[CLS]} token 
as the representation of the 
sentence. Then we compute the cosine similarities between the representations of the input test sentence and  
all source sentences in the training data.

For comparison, we also experiment on single-stage BM25 and BERT representation selection, which serve as baselines in Section \ref{sec:exp}. 

\paragraph{Stage 2: Ungrammatical-syntax-based Ranking} Further, 
we employ \textit{ranking} via syntactic similarity computing  
with Tree Kernel or Polynomial Distance, to obtain the best $k$ matching examples from the candidate set. 

\section{Experimental Results}
\label{sec:exp}

\subsection{Datasets and Evaluation Metrics}
We carry out experiments on English GEC datasets.
Since no model training is involved, most large-scale GEC data is unnecessary, while the data quality matters for example selection.
Thus in this work, we only use the relatively small but high-quality Write\&Improve+LOCNESS (W\&I+LOCNESS) \citep{bryant-etal-2019-bea} as the training data.

For evaluation, we report P (Precision), R (Recall) and F$_{0.5}$ results on BEA-19 test set \citep{bryant-etal-2019-bea} evaluated by ERRANT \citep{bryant-etal-2017-automatic} and on CoNLL-14 test set \citep{ng-etal-2014-conll} evaluated by M2Scorer \citep{dahlmeier-ng-2012-better}. We primarily compare the F$_{0.5}$ among different methods, which shows the comprehensive performance of models on GEC.

Statistics of datasets mentioned above are shown in Table \ref{tab:dataset}.

\begin{table}[H]
\scalebox{0.75}{
\begin{tabular}{lccc}
    \toprule
    \textbf{Dataset} & \textbf{\#Sentences} & \textbf{\%Error} & \textbf{Usage} \\
    \midrule
    W\&I+LOCNESS & 34,308 & 66    & Demonstration \\
    \midrule
    BEA-19-Test & 4,477 & -     & Testing \\
    CoNLL-14-Test & 1,312 & 72    & Testing \\
    \bottomrule
    \end{tabular}%
}
\caption{Statistics of GEC datasets used in this work. \textbf{\#Sentences} refers to the number of sentences.\textbf{\%Error} refers to the percentage of erroneous sentences.}
\label{tab:dataset}
\end{table}
\renewcommand{\thetable}{\arabic{table}}

\subsection{Large Language Models}
\label{sec:models}
We use two mainstream LLM series: LLaMA-2 \citep{touvron2023llama} and GPT-3.5 \citep{openai2023gpt35turbo} for experiment. 

For LLaMA-2, we use \texttt{llama-2-7b-chat} and \texttt{llama-2-13b-chat} with 7B and 13B parameters respectively. For GPT-3.5, we use the official \texttt{gpt-3.5-turbo} API for inference. 

For the sake of reproductivity, we turn off the sampling and set the temperature to zero for all these models we use.



\subsection{Results}
\begin{table*}[htbp]
  \centering
    \scalebox{0.78}{
    \tabcolsep=4pt
    \begin{tabular}{cccccccccccccccccccc}
    \toprule
    \multirow{3}[6]{*}{\textbf{I}} & \multirow{3}[6]{*}{\textbf{II}} & \multicolumn{9}{c}{\textbf{BEA-2019}}                                 & \multicolumn{9}{c}{\textbf{CoNLL-2014}} \\
\cmidrule(lr){3-11}\cmidrule(lr){12-20}          &       & \multicolumn{3}{c}{\textbf{LLaMA-2-7B}} & \multicolumn{3}{c}{\textbf{LLaMA-2-13B}} & \multicolumn{3}{c}{\textbf{GPT-3.5-turbo}} & \multicolumn{3}{c}{\textbf{LLaMA-2-7B}} & \multicolumn{3}{c}{\textbf{LLaMA-2-13B}} & \multicolumn{3}{c}{\textbf{GPT-3.5-turbo}} \\
\cmidrule(lr){3-5}\cmidrule(lr){6-8}\cmidrule(lr){9-11}\cmidrule(lr){12-14}\cmidrule(lr){15-17}\cmidrule(lr){18-20}          &       & \textbf{P} & \textbf{R} & \textbf{F$_{0.5}$} & \textbf{P} & \textbf{R} & \textbf{F$_{0.5}$} & \textbf{P} & \textbf{R} & \textbf{F$_{0.5}$} & \textbf{P} & \textbf{R} & \textbf{F$_{0.5}$} & \textbf{P} & \textbf{R} & \textbf{F$_{0.5}$} & \textbf{P} & \textbf{R} & \textbf{F$_{0.5}$} \\
    \midrule
    \multirow{6}[2]{*}{-} & Rand. & 50.1  & 57.7  & 51.5  & 49.0  & 61.2  & 51.0  & 47.0  & 70.4  & 50.3  & 59.4  & 48.8  & 56.9  & 58.6  & 51.3  & 57.0  & 56.5  & 59.9  & 57.1 \\
          & BM25  & 50.9  & 58.2  & 52.2  & 51.6  & 61.1  & 53.3  & 46.8  & 69.6  & 50.1  & 59.7  & 47.7  & 56.8  & 59.3  & 50.1  & 57.2  & 56.6  & 60.8  & 57.4 \\
          & BERT  & 50.7  & 56.8  & 51.8  & 51.0  & 61.2  & 52.8  & 47.6  & 70.0  & 50.9  & 58.6  & 45.4  & 55.4  & 60.1  & 52.0  & 58.3  & 56.0  & 60.8  & 56.9 \\
          \cdashline{2-20}
          & T. K. & 50.0  & 57.0  & 51.2  & 52.5  & 59.0  & 53.6  & 47.2  & 69.8  & 50.5  & 57.9  & 47.5  & 55.5  & 61.8  & 48.0  & 58.5  & 57.3  & 60.3  & 57.9 \\
          & Poly. & 53.1  & 57.9  & 54.0  & 52.9  & 60.2  & 54.3  & 49.5  & 70.0  & 52.6  & 59.5  & 49.5  & 57.2  & 61.7  & 51.8  & 59.4  & 58.2  & 59.9  & 58.6 \\
          & W. Poly. & 53.2  & 58.2  & \textbf{54.2} & 53.4  & 60.5  & \textbf{54.7} & 50.3  & 69.6  & \uline{\textbf{53.2}} & 60.1  & 49.2  & \textbf{57.5} & 61.6  & 52.3  & \uline{\textbf{59.5}} & 58.4  & 60.5  & \uline{\textbf{58.8}} \\
    \midrule
    \multirow{3}[2]{*}{BM25} & T. K. & 55.1  & 55.9  & \uline{\textbf{55.2}} & 54.9  & 58.7  & \uline{\textbf{55.6}} & 49.7  & 69.3  & \textbf{52.7} & 62.2  & 45.7  & 58.0  & 61.9  & 47.3  & 58.3  & 58.3  & 59.7  & \textbf{58.6} \\
          & Poly. & 51.2  & 57.1  & 52.3  & 50.9  & 59.8  & 52.5  & 48.8  & 69.5  & 51.9  & 62.1  & 47.7  & \uline{\textbf{58.6}} & 60.9  & 49.8  & 58.3  & 57.2  & 59.7  & 57.7 \\
          & W. Poly. & 54.4  & 57.4  & 55.0  & 54.0  & 59.7  & 55.0  & 49.3  & 69.8  & 52.4  & 61.4  & 47.7  & 58.1  & 60.8  & 50.4  & \textbf{58.4} & 57.6  & 60.4  & 58.1 \\
    \midrule
    \multirow{3}[2]{*}{BERT} & T. K. & 53.6  & 56.0  & 54.1  & 53.7  & 59.3  & 54.7  & 50.0  & 69.7  & \textbf{53.0} & 60.7  & 46.3  & 57.1  & 60.8  & 49.9  & \textbf{58.3} & 57.6  & 59.2  & 57.9 \\
          & Poly. & 53.3  & 57.2  & 54.0  & 53.8  & 60.4  & 55.0  & 49.0  & 69.5  & 52.1  & 60.5  & 47.6  & 57.4  & 59.8  & 50.8  & 57.8  & 57.6  & 60.7  & \textbf{58.2} \\
          & W. Poly. & 53.8  & 57.4  & \textbf{54.5} & 54.2  & 60.7  & \textbf{55.4} & 49.9  & 69.7  & 52.9  & 61.0  & 48.3  & \textbf{57.9} & 59.8  & 51.5  & 57.9  & 57.3  & 60.5  & 57.9 \\
    \bottomrule
    \end{tabular}
    }
    \caption{Experimental results under the in-context few-shot setting with 4 examples. 
    \textbf{\MakeUppercase{\romannumeral 1}} and \textbf{\MakeUppercase{\romannumeral 2}} denote the first (\textit{selection}) and second (\textit{ranking}) stage of the two-stage selection respectively. "-" means the Stage \textbf{\MakeUppercase{\romannumeral 1}} is absent and these are single-stage models. "Rand.", "T. K.", "Poly." and "W. Poly." refer to "Random", "Tree Kernel" "Polynomial Distance" and "Weighted Polynomial Distance", respectively. The dashed line separates results of conventional baselines and our proposed methods: the former on the upper side and the latter on the lower side. The best F$_{0.5}$ scores of each 
    group
    are displayed in \textbf{bold}, and the best F$_{0.5}$ scores of all settings are displayed in \uline{\textbf{underlined bold}}.}
  \label{tab:result}
\end{table*}
\renewcommand{\thetable}{\arabic{table}}

Experimental results are shown in Table \ref{tab:result}.
With different LLMs and on both datasets, our ungrammatical-syntax-based selection strategy obviously 
outperforms 
conventional methods. 
On BEA-2019 data, the method with first BM25 selection and then Tree Kernel ranking improves the performance by 3.7, 4.6 and 2.4 F$_{0.5}$ 
points, using \texttt{llama-2-7b-chat}, \texttt{llama-2-13b-chat} and \texttt{gpt-3.5-turbo} respectively.  



\paragraph{Performance of Tree Kernel}
When applied as a single-stage method, the Tree Kernel similarity performs poorly and even achieves a lower F$_{0.5}$ score than conventional baselines. However, with the help of a preliminary 
\textit{selection} stage, it improves by a margin of about 2 to 3 percentage points, and even achieves the highest F$_{0.5}$ score on BEA-2019 data with LLaMA-2.

\paragraph{Performance of Polynomial Distance}
Different from Tree Kernel, Polynomial Distance performs fairly well even without a preliminary \textit{selection}. 
Among those single-stage approaches, both polynomial-based methods outperform 
traditional baselines by an average margin of 2 to 3 percentage points in all cases, which indicates the superiority of syntactic similarity on GEC.
The weighted version, with a higher weight on labels with error
tags, brings a slight improvement in most cases, which shows the effectiveness of error information in GOPar-based dependency trees.

\paragraph{Performance of Two-stage Selection}
As for Tree Kernel, the two-stage selection strategy consistently boosts performance, whether using BM25 or BERT representation as the preliminary selection approach. But for Polynomial Distance, the two-stage selection brings less improvement and even 
fails to improve performance in some cases. We leave it for future research. 

\begin{table}[tbp]
  \centering
  \scalebox{0.65}{
    \tabcolsep=4pt
    \begin{tabular}{lcc}
    \toprule
    \textbf{System} & \textbf{CoNLL-14} & \textbf{BEA-19} \\
    \toprule
    GECToR \citep{omelianchuk-etal-2020-gector} & 65.3  & 72.4 \\
    SynGEC \citep{zhang-etal-2022-syngec} & 66.7 & 72.0 \\
    T5 xxl \citep{rothe-etal-2021-simple} & \textbf{68.9} & \textbf{75.9} \\
    \midrule
    ChatGPT zero-shot CoT \citep{fang2023chatgpt} & 51.7 & 36.1 \\
    \midrule
    LLaMA-2-7B with BM25 + Tree Kernel (ours) & 55.2 & 58.0 \\
    LLaMA-2-13B with Weighed Polynomial (ours) & 54.7 & 59.5 \\
    \bottomrule
    \end{tabular}
    }
  \caption{Results of state-of-the-art GEC systems and our proposed methods on two datasets. 
  The evaluation metric is F$_{0.5}$.}
  \label{tab:sota}%
\end{table}%

\paragraph{Comparison with State-of-the-art}
We compare our method with previous supervised approaches, as shown in Table \ref{tab:sota}. Even with the help of ungrammatical-syntax-based selection, the GEC performance of LLMs is still far from state-of-the-art. 
We look forward to more advanced foundation models in the future.

\section{Model Analysis}
\subsection{Experiments with Different Numbers of In-context Examples}
\label{sec:shot}

\begin{table*}[!t]
  \centering
  \scalebox{0.8}{
    \begin{tabular}{cccccccccccccc}
    \toprule
    \multirow{2}[4]{*}{\textbf{I}} & \multirow{2}[4]{*}{\textbf{II}} & \multicolumn{3}{c}{\textbf{1-shot}} & \multicolumn{3}{c}{\textbf{2-shot}} & \multicolumn{3}{c}{\textbf{4-shot}} & \multicolumn{3}{c}{\textbf{8-shot}} \\
\cmidrule(lr){3-5}\cmidrule(lr){6-8}\cmidrule(lr){9-11}\cmidrule(lr){12-14}          &       & \textbf{P} & \textbf{R} & \textbf{F$_{0.5}$} & \textbf{P} & \textbf{R} & \textbf{F$_{0.5}$} & \textbf{P} & \textbf{R} & \textbf{F$_{0.5}$} & \textbf{P} & \textbf{R} & \textbf{F$_{0.5}$} \\
    \midrule
    \multirow{6}[2]{*}{-} & Rand. & 47.3  & 29.8  & 42.3  & 49.6  & 50.9  & 49.8  & 50.1  & 57.7  & 51.5  & 52.2  & 58.8  & 53.4  \\
          & BM25  & 48.4  & 35.8  & \textbf{45.2} & 50.4  & 53.1  & 50.9  & 50.9  & 58.2  & 52.2  & 52.5  & 59.0  & 53.7  \\
          & BERT  & 47.3  & 33.8  & 43.8  & 50.0  & 51.4  & 50.2  & 50.7  & 56.8  & 51.8  & 53.6  & 59.3  & 54.6  \\
          \cdashline{2-14}
          & T. K. & 47.1  & 27.4  & 41.2  & 49.0  & 53.2  & 49.8  & 50.0  & 57.0  & 51.2  & 53.6  & 55.9  & 54.0  \\
          & Poly. & 50.1  & 31.5  & 44.8  & 53.9  & 51.9  & \uline{\textbf{53.5}} & 53.1  & 57.9  & 54.0  & 54.3  & 58.3  & \textbf{55.1} \\
          & W. Poly. & 50.4  & 31.5  & 45.0  & 52.7  & 51.5  & 52.4  & 53.2  & 58.2  & \textbf{54.2} & 53.3  & 58.0  & 54.2  \\
    \midrule
    \multirow{3}[2]{*}{BM25} & T. K. & 51.7  & 37.5  & \uline{\textbf{48.1}} & 53.3  & 53.8  & \textbf{53.4} & 55.1  & 55.9  & \uline{\textbf{55.2}} & 57.2  & 55.6  & \textbf{56.9} \\
          & Poly. & 51.3  & 36.6  & 47.5  & 52.9  & 54.5  & 53.2  & 51.2  & 57.1  & 52.3  & 55.5  & 56.9  & 55.8  \\
          & W. Poly. & 51.1  & 36.6  & 47.4  & 52.8  & 54.7  & 53.2  & 54.4  & 57.4  & 55.0  & 56.3  & 57.0  & 56.4  \\
    \midrule
    \multirow{3}[2]{*}{BERT} & T. K. & 50.7  & 35.6  & 46.8  & 53.3  & 52.4  & \textbf{53.1} & 53.6  & 56.0  & 54.1  & 57.1  & 57.0  & \uline{\textbf{57.1}} \\
          & Poly. & 50.9  & 35.5  & \textbf{46.9} & 52.1  & 53.4  & 52.4  & 53.3  & 57.2  & 54.0  & 55.5  & 58.2  & 56.1  \\
          & W. Poly. & 50.6  & 35.7  & 46.7  & 52.1  & 53.8  & 52.4  & 53.8  & 57.4  & \textbf{54.5} & 56.5  & 57.8  & 56.7  \\
    \bottomrule
    \end{tabular}
  }
  \caption{Results of \texttt{llama-2-7b-chat} with different numbers of shots on BEA-19 test set.}
  \label{tab:bea19shot}
\end{table*}

\begin{table*}[!t]
  \centering
  \scalebox{0.8}{
    \begin{tabular}{cccccccccccccc}
    \toprule
    \multirow{2}[4]{*}{\textbf{I}} & \multirow{2}[4]{*}{\textbf{II}} & \multicolumn{3}{c}{\textbf{1-shot}} & \multicolumn{3}{c}{\textbf{2-shot}} & \multicolumn{3}{c}{\textbf{4-shot}} & \multicolumn{3}{c}{\textbf{8-shot}} \\
\cmidrule(lr){3-5}\cmidrule(lr){6-8}\cmidrule(lr){9-11}\cmidrule(lr){12-14}          &       & \textbf{P} & \textbf{R} & \textbf{F$_{0.5}$} & \textbf{P} & \textbf{R} & \textbf{F$_{0.5}$} & \textbf{P} & \textbf{R} & \textbf{F$_{0.5}$} & \textbf{P} & \textbf{R} & \textbf{F$_{0.5}$} \\
    \midrule
    \multirow{6}[2]{*}{-} & Rand. & 54.7  & 21.2  & 41.6  & 58.0  & 42.3  & 54.0  & 59.1  & 48.2  & 56.6  & 60.9  & 49.3  & 58.2  \\
          & BM25  & 55.7  & 25.2  & \textbf{44.9} & 57.5  & 42.5  & 53.7  & 59.7  & 47.7  & 56.8  & 60.4  & 47.8  & 57.4  \\
          & BERT  & 55.9  & 22.5  & 43.1  & 58.0  & 39.0  & 52.8  & 58.6  & 45.4  & 55.4  & 60.7  & 48.0  & 57.6  \\
          \cdashline{2-14}
          & T. K. & 51.5  & 17.7  & 37.3  & 57.9  & 44.1  & 54.5  & 57.9  & 47.5  & 55.5  & 61.7  & 47.0  & 58.1  \\
          & Poly. & 54.7  & 21.3  & 41.6  & 58.6  & 41.4  & 54.1  & 59.6  & 49.5  & 57.2  & 60.5  & 48.5  & 57.6  \\
          & W. Poly. & 52.3  & 20.6  & 40.0  & 58.6  & 42.9  & \textbf{54.6} & 60.1  & 49.2  & \textbf{57.5} & 61.0  & 49.8  & \textbf{58.4} \\
    \midrule
    \multirow{3}[2]{*}{BM25} & T. K. & 57.9  & 27.3  & \uline{\textbf{47.3}} & 60.5  & 44.7  & \uline{\textbf{56.5}} & 62.2  & 45.7  & 58.0  & 62.5  & 45.3  & 58.1  \\
          & Poly. & 57.2  & 25.1  & 45.5  & 60.5  & 43.5  & 56.2  & 62.1  & 47.7  & \uline{\textbf{58.6}} & 61.6  & 46.7  & 57.9  \\
          & W. Poly. & 57.1  & 24.6  & 45.1  & 60.7  & 43.7  & 56.3  & 61.4  & 47.7  & 58.1  & 62.7  & 47.7  & \textbf{59.0} \\
    \midrule
    \multirow{3}[2]{*}{BERT} & T. K. & 58.3  & 25.1  & \textbf{46.1} & 59.9  & 42.7  & 55.4  & 60.7  & 46.3  & 57.1  & 63.1  & 46.2  & 58.8  \\
          & Poly. & 56.0  & 24.7  & 44.7  & 59.3  & 43.8  & 55.4  & 60.5  & 47.6  & 57.4  & 61.9  & 47.7  & 58.4  \\
          & W. Poly. & 57.1  & 24.9  & 45.4  & 59.5  & 44.6  & \textbf{55.8} & 61.0  & 48.3  & \textbf{57.9} & 62.8  & 47.8  & \uline{\textbf{59.1}} \\
    \bottomrule
    \end{tabular}%
  }
  \caption{Results of \texttt{llama-2-7b-chat} with different numbers of shots on CoNLL-14 test set.}
  \label{tab:conll14shot}%
\end{table*}%

To explore the consistency and robustness of our methods, we conduct 1-shot, 2-shot, 4-shot and 8-shot experiments on \texttt{llama-2-7b-chat}.
The results on BEA-2019 test set and CoNLL-2014 test set are shown in Table \ref{tab:bea19shot} and \ref{tab:conll14shot} respectively.

When there is only one example, the model performs relatively poor. When the number of examples comes to two, the performance improves significantly. Then, further increasing 
the number of examples brings a slight but consistent performance gain. 

When the number of examples is small, the superiority of syntax-based methods compared with those conventional is evident. When the number of examples increases, conventional baselines improve a lot while syntax-based methods gain relatively less, which shows a marginal benefit. But 
syntax-based methods always secure the highest score, indicating the consistency of their advantages, especially under settings of fewer shots.

\subsection{Ungrammatical Parser or Standard Parser?}

\definecolor{brickred}{HTML}{b92622}
\definecolor{midnightblue}{HTML}{005c7f}
\definecolor{salmon}{HTML}{f1958d}
\definecolor{burntorange}{HTML}{f19249}
\definecolor{junglegreen}{HTML}{4dae9d}
\definecolor{forestgreen}{HTML}{499c5e}
\definecolor{pinegreen}{HTML}{3d8a75}
\definecolor{seagreen}{HTML}{6bc1a2}
\definecolor{limegreen}{HTML}{97c65a}

\begin{figure}[htbp]
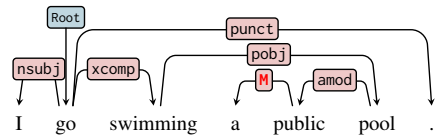
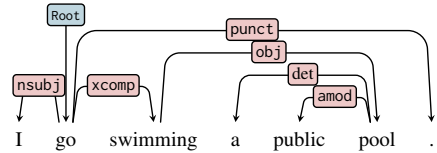

  \begin{subfigure}[b]{0.5\textwidth}
    \centering
    \scalebox{0.65}{
        \begin{dependency}
          \begin{deptext}[column sep=0.5cm,font=\small]
            \large{I} \& \large{go} \& \large{swimming} \& \large{a} \& \large{public} \& \large{pool} \& \large{.} \\ 
          \end{deptext}
          \Large{
          \deproot[edge vertical padding=0.6ex, edge height=9.5ex, label style={fill=midnightblue!25, thick}, edge style={thick}]{2}{\texttt{\large{Root}}}
          \depedge[edge vertical padding=0.6ex, edge height=3.2ex, label style={fill=brickred!25, thick}, thick]{2}{1}{\texttt{nsubj}}
          \depedge[edge vertical padding=0.6ex, edge height=3.2ex, label style={fill=brickred!25, thick}, thick]{2}{3}{\texttt{xcomp}}
            \depedge[edge vertical padding=0.6ex, edge height=4.5ex, label style={fill=brickred!25, thick}, thick]{3}{6}{\texttt{pobj}}
          \depedge[edge vertical padding=0.6ex, edge height=6.8ex, label style={fill=brickred!25, thick}, thick]{2}{7}{\texttt{punct}}
          \depedge[edge vertical padding=0.6ex, edge height=2.2ex, label style={fill=brickred!25, thick}, thick]{5}{4}{\textcolor{red}{\texttt{\textbf{M}}}}
          \depedge[edge vertical padding=0.6ex, edge height=2.2ex, label style={fill=brickred!25, thick}, thick]{6}{5}{\texttt{amod}}}
    
        \end{dependency}
    }
    
    \caption{GOPar.}
    \label{fig:gopar}
  \end{subfigure}
  \\
   \begin{subfigure}[b]{0.5\textwidth}
    \centering
    \scalebox{0.65}{
    
    \begin{dependency}
          \begin{deptext}[column sep=0.5cm,font=\small]
            \large{I} \& \large{go} \& \large{swimming} \& \large{a} \& \large{public} \& \large{pool} \& \large{.} \\ 
          \end{deptext}
          \Large{
          \deproot[edge vertical padding=0.6ex, edge height=11.2ex, label style={fill=midnightblue!25, thick}, edge style={thick}]{2}{\texttt{\large{Root}}}
          \depedge[edge vertical padding=0.6ex, edge height=3.2ex, label style={fill=brickred!25, thick}, thick]{2}{1}{\texttt{nsubj}}
          \depedge[edge vertical padding=0.6ex, edge height=3.2ex, label style={fill=brickred!25, thick}, thick]{2}{3}{\texttt{xcomp}}
            \depedge[edge vertical padding=0.6ex, edge height=6.2ex, label style={fill=brickred!25, thick}, thick]{3}{6}{\texttt{obj}}
          \depedge[edge vertical padding=0.6ex, edge height=8.2ex, label style={fill=brickred!25, thick}, thick]{2}{7}{\texttt{punct}}
          \depedge[edge vertical padding=0.6ex, edge height=4.2ex, label style={fill=brickred!25, thick}, thick]{6}{4}{det}
          \depedge[edge vertical padding=0.6ex, edge height=2.2ex, label style={fill=brickred!25, thick}, thick]{6}{5}{\texttt{amod}}}
    
        \end{dependency}
    }
    \caption{Stanford Parser.}
    \label{fig:stanford}
  \end{subfigure}
  \hfill
  \hfill
  \caption{An example of parsing tree by
  GOPar and Stanford Parser.}
  \label{fig:goparvsstanford}
\end{figure}

\begin{table*}[!t]
  \centering
    \scalebox{0.7}{
    \begin{tabular}{c|l|l}
    \hline
          & \textbf{Source (Erroneous Sentence)} & \textbf{Target (Corrected Sentence)} \\
    \hline
    Input & So, they 
    \textit{have to also} prepare mentally. & So, they 
    \textit{also have to} prepare mentally. \\
    \hline
    BM25  & Also you can see how they prepare your food in front of you. & Also
    \textit{,} you can see how they prepare your food in front of you. \\
    T. K. & Nowadays people 
    \textit{get around constantly}. & Nowadays
    \textit{,} people 
    \textit{are constantly on the move}. \\
    BM25 + T. K. & 
    \textit{that have limitation also} there. & 
    \textit{There are also limitations} there. \\
    \hline
    \end{tabular}
    }
  \caption{A one-shot example showing the tree kernel method benefiting from the two-stage selection.}
  \label{tab:case}
\end{table*}
\renewcommand{\thetable}{\arabic{table}}

\begin{table}[htbp]
  \centering
    \scalebox{0.8}{
    \begin{tabular}{cccccccc}
    \toprule
    \multirow{2}[4]{*}{\textbf{\MakeUppercase{\romannumeral 1}}} & \multirow{2}[4]{*}{\textbf{\MakeUppercase{\romannumeral 2}}} & \multicolumn{3}{c}{\textbf{GOPar}} & \multicolumn{3}{c}{\textbf{Stanford Parser}} \\
\cmidrule(lr){3-5} \cmidrule(lr){6-8}          &       & \textbf{P}     & \textbf{R}     & \textbf{F$_{0.5}$}     & \textbf{P}     & \textbf{R}     & \textbf{F$_{0.5}$} \\
    \midrule
    -     & \multirow{3}[2]{*}{T. K.} & 50.0  & 57.0  & 51.2  & 49.6  & 56.4  & 50.8 \\
    BM25  &       & 55.1  & 55.9  & \textbf{55.2} & 51.8  & 56.2  & \textbf{52.7} \\
    BERT  &       & 53.6  & 56.0  & 54.1  & 50.6  & 57.2  & 51.8 \\
    \bottomrule
    \end{tabular}
    }
  \caption{Results on BEA-2019 test set with 4 examples, using 
  GOPar and Stanford Parser respectively.}
  \label{tab:stanford}
\end{table}%
\renewcommand{\thetable}{\arabic{table}}

To explore the affect of different parsers on model performance, 
we also experiment with Stanford Parser \citep{dozat2017deep}, which is a widely-used conventional 
parser. 
We use \texttt{stanford-corenlp-4.5.5} as the model for parsing and run Stanford Parser with \texttt{stanfordcorenlp}~\footnote{https://pypi.org/project/stanfordcorenlp}, which is a Python wrapper for Stanford CoreNLP.
For a clear demonstration, an example is illustrated in Figure \ref{fig:goparvsstanford} to show the different parsing results of GOPar and Stanford Parser.

The experimental results 
comparing GOPar and Stanford Parser on BEA-2019 test set are shown in Table \ref{tab:stanford}. Here, we adopt \texttt{llama-2-7b-chat} as the LLM and Tree Kernel as the ranking method.


Without using the two-stage selection, Stanford Parser performs slightly worse than GOPar. With the two-stage selection, GOPar gains more improvement than Stanford Parser and outperforms it by a margin of more than 2 points. This indicates GOPar is more suitable for GEC, and its superiority lies in two aspects. First, it performs more robust on ungrammatical sentences (e.g., it correctly recognizes the prepositional object 
"pool" in the sentence shown in Figure \ref{fig:goparvsstanford} while Stanford Parser fails to). Second, it provides extra information about the grammatical errors (e.g., the \textit{Missing} error in Figure \ref{fig:goparvsstanford}).

\subsection{Effect of Two-stage Selection}

In order to find out how the two-stage strategy benefits the Tree Kernel method, we conduct a case study and compare three selection settings: BM25 only ("BM25"), 
Tree Kernel only ("T.K.") 
and Tree Kernel after 
the BM25 \textit{selection} ("BM25+T.K."). 

In the example shown in Table \ref{tab:case}, the input sentence is ungrammatical in word order. "BM25" selects a sentence with a punctuation missing error that is similar to the input sample in words ("also", "they" and "prepare"). "T. K." selects a sentence with an improper expression "get around constantly" which is similar to "prepare mentally" 
in syntactic structure but has little to do with the grammatical errors. "BM25 + T. K." selects a sentence that is similar to the input sample both in word occurrences ("also" and "have") and in error form (improper word order).

Since similar words are more likely to form similar errors, with the help of a preliminary selection, Tree Kernel can select from a more relative candidate set, 
leading to a better 
example selection involving both word and syntactic similarity in erroneous constituents.  
Moreover, it also shows the disadvantage of conventional selection method BM25 on GEC, which 
cannot effectively select examples similar in syntax.

\section{Conclusion}
In this work, we make use of two conventional tree-based syntactic similarity algorithms and the select-then-rank two-stage framework to select in-context examples for the GEC task. Empirical results show that our syntax-based in-context example selection method is effective for GEC. We call on the NLP community to pay more attention to the help of syntactic information for many other syntax-related tasks besides GEC.

\section*{Acknowledgements}
This work is supported by the National Natural Science Foundation of China (62076008) and the Key Project of Natural Science Foundation of China (61936012).


\section*{Limitations}
First, we only experiment on English datasets. The performance of our method on other languages requires further exploration. Second, besides dependency tree, constituent tree is also worth trying. However, unfortunately, we do not have access to GEC-oriented constituent trees \citep{zhang2022csyngec} at the time of writing this paper. Third, many previous outstanding methods of both in-context example selection and tree similarity computation have not been explored in our work. Fourth, due to limited time, we do not explore the effect of the size of candidate set after the \textit{selection} stage and the choice of weight of ungrammatical nodes in the Polynomial Distance method. There may exist a better size than the values we use in our experiments. Fifth, except for the Stanford Parser (which splits sentences itself), our experiments do not split instances with multiple sentences into single-sentence instances. Some instances in GEC datasets contains more than one sentence. Directly parsing these instances without splitting them into single sentences may hurt the parsing performance and lead to unreliable results. Last, we do not treat in-context examples as a whole, which might lead to a lower level of diversity of examples and sub-optimal performance, as addressed in \citet{pmlr-v202-ye23c} and \citet{gupta-etal-2023-coverage}.

\section*{Ethics Statement}
\paragraph{Use of Scientific Artifacts.}
We make use of GOPar provided by \citet{zhang-etal-2022-syngec}, which is publicly available based on the MIT license~\footnote{https://github.com/HillZhang1999/SynGEC}.

\paragraph{About Computational Budget.}
Computation time is shown in Table \ref{tab:time}.
\begin{table}[htbp]
  \centering
  \scalebox{0.8}{
    \tabcolsep=4pt
    \begin{tabular}{cc}
    \toprule
    \textbf{Method} & \textbf{Time} \\
    \midrule
    BM25  & 440 \\
    BERT  & 4500 \\
    Tree Kernel & 3600 \\
    Polynomial Distance & 3200 \\
    \bottomrule
    \end{tabular}%
    }
  \caption{Computation time of different methods on BEA-19 test set, all in seconds. BERT runs on an NVIDIA GeForce RTX 2080 Ti and the other three run on an Intel$^\circledR$ Xeon$^\circledR$ Gold 5218 CPU.}
  \label{tab:time}
\end{table}%

\paragraph{About Reproducibility.}
All the experiments are completely reproducible since we disable sampling and set the temperature to zero for all LLMs we use, as discussed in Section \ref{sec:models}.

\normalem
\bibliography{anthology,custom}









\end{document}